\PassOptionsToClass{balance=false}{acmart}
\documentclass[sigconf,nonacm]{acmart}
\usepackage{amsmath,booktabs,tabularx,array,multirow,pifont,colortbl}
\usepackage{enumitem,xspace}
\usepackage{xcolor}
\acmYear{2027}
\copyrightyear{2027}
\usepackage{algorithm}
\usepackage{algpseudocode}

\setcopyright{none}
\acmDOI{}
\acmISBN{}
\definecolor{sdblue}{HTML}{205B91}
\definecolor{sdgreen}{HTML}{16723E}
\definecolor{sdred}{HTML}{B3242B}
\definecolor{sdshade}{HTML}{DDEFE2}
\definecolor{sdgood}{HTML}{BFE1C9}
\newcommand{\yes}{\textcolor{sdgreen}{\ding{51}}}
\newcommand{\no}{\textcolor{sdred}{\ding{55}}}

\newcommand{\ours}{\textsc{Skill Dreaming}\xspace}

\newcolumntype{L}[1]{>{\raggedright\arraybackslash}p{#1}}
\newcolumntype{Y}{>{\raggedright\arraybackslash}X}
\setlist[itemize]{leftmargin=*,itemsep=2pt,topsep=3pt}
\AtEndPreamble{%
  \hypersetup{%
    colorlinks=true,
    allcolors=blue,
    citecolor=black,
    citebordercolor={0 0.65 0}%
  }%
}
\makeatletter
\AtBeginDocument{%
  \hypersetup{pdfborder={0 0 0},pdfborderstyle={/S /S /W 0}}%
  \pretocmd{\hyper@natlinkstart}{%
    \begingroup
    \hypersetup{pdfborder={0 0 0.5},pdfborderstyle={/S /S /W 0.5}}%
  }{}{\PackageError{SkillDreaming}{Cannot style citation links}{Check hyperref and natbib compatibility.}}%
  \apptocmd{\hyper@natlinkend}{\endgroup}{}%
    {\PackageError{SkillDreaming}{Cannot close citation styling}{Check hyperref and natbib compatibility.}}%
}
\makeatother
\begin{document}

\title{\ours: Evolving Repository Skills through Pull Request Streams}
\author{Xinyuan Song}
\affiliation{\institution{Causal Dynamics Lab}\country{USA}}

\author{Hasibul Haque}
\affiliation{\institution{Causal Dynamics Lab}\country{USA}}

\author{Jay Somasundaram}
\affiliation{\institution{Causal Dynamics Lab}\country{USA}}

\author{Ryan Turner}
\affiliation{\institution{Causal Dynamics Lab}\country{USA}}

\author{Bowen Zhu}
\affiliation{\institution{Causal Dynamics Lab}\country{USA}}

\author{Liang Zhao}
\affiliation{\institution{Causal Dynamics Lab}\country{USA}}
\renewcommand{\shortauthors}{Song et al.}

\begin{abstract}

Large language model (LLM) agents increasingly rely on reusable skills for complex software engineering tasks, making skill libraries essential for repository development and maintenance. However, repository-specific skills must evolve with changing code structures, implementation conventions, and maintenance requirements. Pull requests (PRs) provide a natural source of such knowledge by recording how repositories evolve through implementation and review. Meanwhile, new PRs must follow repository-specific implementation patterns, modify related components consistently, and ensure that existing tests continue to pass while adding coverage for new functionality. This creates a mutually reinforcing cycle: PRs provide knowledge for updating skills, while these skills improve future PRs, maintain repositories, and support further skill evolution. To enable this co-evolution of PRs, repositories, and skills, we propose \ours, a framework that learns from PR streams to guide continuous repository development and maintenance. Given a repository and its PR history, \ours builds a repository-specific skill library by extracting executable knowledge from validated PRs and continuously refines the skills as new PR information becomes available. These skills guide coding agents not only in implementing requested features but also in maintaining associated tests, documentation, compatibility, and repository conventions. By integrating PR analysis, skill management, code execution, and continuous integration, \ours provides an end-to-end workflow for applying learned skills to repository changes and incorporating subsequent PR outcomes into future skill updates. On the SWE-bench Pro benchmark, \ours increases the official resolved rate from 79\% for the base agent to 88\%. Additional experiments demonstrate the effectiveness of \ours across different repositories, highlighting the potential of PR-driven skill evolution for reliable software development.

\end{abstract}


\maketitle

\section{Introduction}

\begin{figure*}[!ht]
\centering
\includegraphics[width=.9\textwidth]{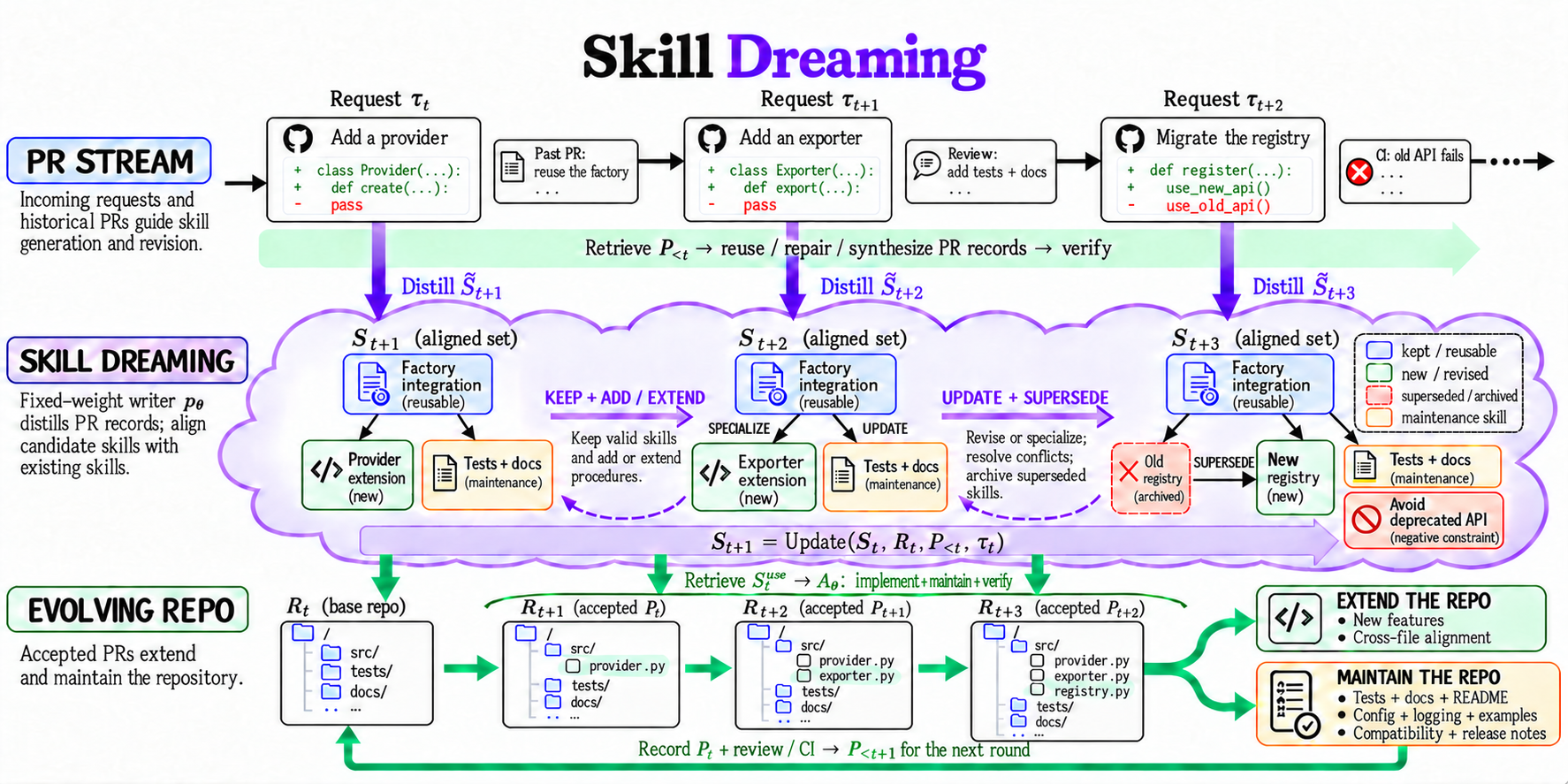}
\caption{
\textbf{Overview of \ours: PR-driven skill evolution.}
\textbf{Top:} PR streams provide development records.
\textbf{Middle:} Skills evolve through generation, reuse, revision, and replacement.
\textbf{Bottom:} Evolving skills guide repository development and maintenance, generating feedback for further evolution.
}
\Description{
Three interconnected layers illustrate the continuous evolution of PR streams, repository-specific skills, and software repositories.
}
\label{fig:story}
\end{figure*}

Large language model (LLM) agents increasingly rely on reusable skills to perform complex software engineering tasks~\cite{cao2026skillmarket,tong2026code2skill,gong2026skillmoo,ling2026agentskills}. By capturing knowledge from previous tasks, skills allow agents to reuse successful strategies, reduce repeated exploration, and improve performance on future tasks. Existing approaches have explored different mechanisms for skill learning and reuse~\cite{zhou2026agentskillssurvey,pu2026skillopsmanagingllmagent}. Trace2Skill extracts transferable lessons from execution trajectories~\cite{ni2026trace2skill}, while ReasoningBank organizes successful and unsuccessful experiences into reusable reasoning memory~\cite{ouyang2025reasoningbank}. CODESKILL studies self-evolving skill libraries for coding agents~\cite{li2026codeskill}, and Socratic-SWE uses trace-derived skills to support iterative agent improvement~\cite{xiao2026socratic}. SkillLens investigates adaptive skill reuse at different levels of granularity~\cite{yu2026skilllens}, while Learning to Commit explores repository-specific knowledge from chronological development histories~\cite{li2026commit}. However, existing approaches primarily focus on learning and reusing skills from past experiences, with limited support for continuously aligning these skills with evolving repositories. As repository structures, APIs, and implementation conventions change, previously learned skills may become incomplete, outdated, or inconsistent with current development requirements~\cite{li2026dynamicagentskills}. This highlights the need for agents to continuously learn and reuse repository-specific skills from new development records. These skills must also be updated as repositories evolve to remain relevant for future tasks.

Pull requests (PRs) provide a valuable source of repository-specific knowledge. They document how development requests are addressed through code changes~\cite{jimenez2024swebench,deng2025swebenchpro} and how these changes are reviewed and validated~\cite{xu2026agentprs}. This history reveals implementation patterns and conventions that can guide future repository development and maintenance. More importantly, successive PRs provide a history of repository evolution~\cite{li2026commit,li2026featx}. For example, an initial PR may introduce a provider and its factory integration, a later PR may reuse this convention for an exporter, and a subsequent PR may migrate the registry to a new API. Learning from this history allows agents to capture implementation conventions, maintain documentation, and evolve repository-specific skills~\cite{li2026dynamicagentskills} as the codebase changes. Retrieval-augmented generation (RAG)~\cite{lewis2020rag} offers a natural approach to extracting relevant knowledge from historical PRs, allowing agents to retrieve implementation examples and repository context~\cite{zhang2023repocoder,asai2024selfrag}. However, RAG primarily retrieves existing PR information for individual tasks, without learning persistent repository-specific knowledge from development history. As new PRs introduce changes, previously retrieved patterns may become outdated, while new conventions and maintenance requirements remain scattered across the history. This motivates learning reusable skills from PR streams and continuously updating them to guide repository development and maintenance.

To address these challenges, we propose \textbf{\ours}, a framework that learns to extend and maintain repositories from PR streams. As shown in Figure~\ref{fig:story}, \ours establishes a cycle among PR information, repository-specific skills, and repository development. Given a repository and its PR history, the framework extracts reusable knowledge from validated PRs~\cite{tong2026code2skill} into a dynamic skill library~\cite{tang2026wikiskill}. These skills capture implementation procedures, architectural conventions, and maintenance requirements~\cite{gao2026registryskill}, enabling coding agents to apply historical knowledge to new tasks. As agents extend and maintain repositories, resulting PRs provide information for updating the skill library~\cite{wang2026skillevolution}. Thus, PRs drive skill learning, while learned skills guide repository evolution.

The core of \ours is a three-stage framework connecting skill learning, repository development, and skill evolution. First, given an existing repository and its historical PRs, the framework extracts reusable implementation and maintenance knowledge to construct an initial repository-specific skill library. Second, coding agents retrieve relevant skills to guide new development tasks, including feature implementation, testing, documentation, and configuration updates. Third, new PR information from code changes, reviews, and validation drives skill evolution through reuse, extension, revision, and replacement. This closed-loop process is formulated as a sequential optimization problem that jointly maximizes task quality and the future utility of learned skills while minimizing development costs under correctness and safety constraints. Through this process, \ours continuously improves repository-specific knowledge and adapts to evolving development.

We evaluate \ours on repository-level software engineering tasks across three aspects. First, on SWE-bench Pro~\cite{deng2025swebenchpro}, \ours improves the official resolved rate from 79\% with the base agent to 88\%, demonstrating strong performance in issue resolution. Second, cross-repository experiments show consistent improvements over baselines, highlighting its effectiveness across diverse repositories. Third, additional experiments demonstrate the benefits of skill reuse, continuous evolution, and repository maintenance. Together, these results validate the effectiveness of PR-driven skill evolution for repository-level software engineering.

Our main contributions are summarized as follows:
\begin{itemize}
    \item We formulate PR-driven repository skill learning as a continuous process where historical PRs provide reusable knowledge, learned skills guide repository development, and new PRs drive further skill evolution.

    \item We propose \ours, an end-to-end framework that builds repository-specific skill libraries from PR histories, applies learned skills to repository development and maintenance, and continuously evolves them without updating LLM parameters.

    \item We conduct comprehensive experiments demonstrating improved issue resolution on SWE-bench Pro, strong cross-repository performance, and the effectiveness of skill reuse, evolution, and repository maintenance.
\end{itemize}

\section{Related Work}
\label{sec:related}

\begin{table*}[!ht]
\caption{\textbf{Comparison with existing skill-learning and repository-editing methods.} \yes~indicates an explicitly supported mechanism, while \no~indicates otherwise.}
\label{tab:related-matrix}
\centering
\setlength{\tabcolsep}{5pt}
\resizebox{0.95\textwidth}{!}{
\begin{tabular}{l*{7}{>{\centering\arraybackslash}p{1.8cm}}}
\toprule
Method & PR Learning & Skill Library & Skill Reuse & Skill Evolution & Version Awareness & Repo Maintenance & Training-Free \\
\midrule
ReasoningBank~\cite{ouyang2025reasoningbank}
& \no & \yes & \yes & \yes & \no & \no & \yes \\
Trace2Skill~\cite{ni2026trace2skill}
& \no & \yes & \yes & \yes & \no & \no & \yes \\
Learning to Commit~\cite{li2026commit}
& \yes & \yes & \yes & \yes & \no & \yes & \yes \\
CODESKILL~\cite{li2026codeskill}
& \no & \yes & \yes & \yes & \no & \yes & \no \\
Socratic-SWE~\cite{xiao2026socratic}
& \no & \yes & \yes & \no & \no & \yes & \no \\
SkillLens~\cite{yu2026skilllens}
& \no & \yes & \yes & \yes & \no & \no & \yes \\
FeatX~\cite{li2026featx}
& \no & \no & \no & \no & \no & \yes & \yes \\
\rowcolor{blue!10}
\textbf{\ours (ours)}
& \yes & \yes & \yes & \yes & \yes & \yes & \yes \\
\bottomrule
\end{tabular}
}
\end{table*}

\paragraph{Learning repository conventions from history.}
Learning to Commit~\cite{li2026commit} extracts repository-specific development knowledge from historical commits through contrastive reflection, transforming successful implementation patterns and past mistakes into reusable skills. It also supports skill revision and deprecation as repository practices change. Similarly, \ours learns from historical development activities, but uses PR records that incorporate implementation changes, review feedback, and validation outcomes. Rather than treating extracted knowledge as independent instructions, \ours aligns new information with existing skills, resolves conflicting conventions, and preserves version-specific implementation and maintenance requirements. This enables skills to remain applicable as repository structures and development practices evolve.

\paragraph{Experience distillation and skill management.}
Trace2Skill~\cite{ni2026trace2skill} extracts reusable skills from successful and failed agent execution trajectories, allowing agents to transfer learned strategies to future tasks. ReasoningBank~\cite{ouyang2025reasoningbank} organizes reasoning experiences into reusable memories that support subsequent decision-making. CODESKILL~\cite{li2026codeskill} learns a skill manager to extract, organize, and evolve skills from coding experiences. These approaches demonstrate the value of accumulating reusable knowledge, but primarily emphasize agent experiences or learned skill management. In contrast, \ours uses repository development records as its primary evidence source and maintains an evolving skill library without updating LLM parameters. Its skill construction and alignment processes explicitly account for repository conventions, implementation dependencies, and maintenance obligations.

\paragraph{Skill-guided task generation and hierarchical reuse.}
Socratic-SWE~\cite{xiao2026socratic} distills software engineering knowledge into skills and uses them to generate training tasks, connecting reusable experience with task synthesis. SkillLens~\cite{yu2026skilllens} organizes skills hierarchically and supports selective updates, allowing agents to identify and reuse relevant knowledge at different levels of abstraction. While these methods explore how skills can support learning and reuse, \ours focuses on how repository-specific skills should evolve alongside successive PRs. It determines whether incoming PR information should introduce new skills, refine existing procedures, specialize reusable knowledge, or replace outdated instructions. These decisions are informed by repository changes and the relationships among existing skills, allowing the skill library to adapt without indiscriminate accumulation.

\paragraph{Repository editing and PR analysis.}
FeatX~\cite{li2026featx} studies feature-oriented repository editing, addressing the need to coordinate changes across multiple files and components. Agentless~\cite{xia2025agentless} decomposes automated software repair into localization, patch generation, and validation, providing a structured approach to repository-level problem solving. Agent-authored PR analysis~\cite{xu2026agentprs} examines the characteristics and challenges of agent-generated contributions, while SWE-bench and its extensions~\cite{jimenez2024swebench,deng2025swebenchpro,openai2024verified} provide executable benchmarks for evaluating software engineering agents. These studies establish important foundations for repository editing and evaluation. \ours complements them by connecting historical PR information with future implementation: validated development records are transformed into reusable skills that guide feature implementation and associated maintenance, and subsequent PR outcomes provide new evidence for skill evolution.

\paragraph{Comparison with existing methods.}
Table~\ref{tab:related-matrix} compares existing approaches across PR-driven learning, skill evolution, repository maintenance, and training-free adaptation. Prior work has explored repository-specific skill extraction and deprecation~\cite{li2026commit}, experience-based skill distillation~\cite{ni2026trace2skill}, hierarchical skill management~\cite{yu2026skilllens}, and skill-guided task synthesis~\cite{xiao2026socratic}. However, these capabilities are typically studied separately. \ours integrates PR record construction, evidence-grounded skill distillation, skill relationship management, and skill-guided repository implementation into a continuous process. This connection between historical development evidence, evolving skills, and subsequent repository changes distinguishes \ours from existing approaches.

\section{Problem Formulation}
\label{sec:problem}

\paragraph{Repository state and skill set.}
At round $t$, the system maintains a repository $R_t$ and a skill set $S_t$. A pull request (PR) represents a proposed repository change, including its description, code modifications, and associated feedback. Let $P_{<t}=\{P_1,\ldots,P_{t-1}\}$ denote previously available PRs, and let $\tau_t$ denote a new development or maintenance request. The skill set $S_t$ captures repository-specific knowledge learned from previous PRs, including implementation patterns, coding conventions, and maintenance requirements. Given $\tau_t$, the system aims to produce a new PR $P_t$ that implements the requested changes. As the repository evolves, new PR information can inform the reuse, refinement, extension, or replacement of existing skills.
\paragraph{PR information and skill evolution.}
The system first constructs PR records through three operations: detecting relevant historical PRs, repairing incomplete or unsuccessful implementations, and synthesizing new candidate changes. The resulting candidates are verified against repository requirements to identify successful changes and informative failures. Based on this information, the system updates the skill set:
\begin{equation}
\label{eq:update}
S_{t+1}=\operatorname{Update}(S_t,R_t,P_{<t},\tau_t).
\end{equation}
The update operation extracts reusable engineering knowledge from the constructed PR records and incorporates it into existing skills. Skills can be reused, extended, revised, or replaced as repository requirements evolve, producing an updated skill set that guides subsequent development.

The updated skill set $S_{t+1}$ guides a fixed-weight agent $A_\theta$ to generate the new PR and update the repository:
\begin{equation}
\label{eq:execution}
P_t=A_\theta(R_t,\tau_t,S_{t+1}),\quad
R_{t+1}=
\begin{cases}
R_t + P_t, & \text{if accepted},\\
R_t,       & \text{otherwise}.
\end{cases}
\end{equation}
Here, $R_t + P_t$ denotes the repository after incorporating the accepted PR. The generated PR $P_t$ and its feedback become part of the historical records $P_{<t+1}$, informing subsequent skill updates. In historical evaluation, each request $\tau_t$ uses its recorded base repository $R_t$, while only the skill set $S_t$ evolves across requests without chaining generated repository changes.

\paragraph{PR-guided optimization objective.}
The objective is to continuously improve repository development by learning from previous PRs and evolving the skill set across successive requests. Given repository $R_t$, previous PRs $P_{<t}$, and request $\tau_t$, we define the utility of a generated PR as
\begin{equation}
\label{eq:utility}
U_t(P_t,R_t,\tau_t,S_{t+1})=\Pr(P_t \text{ satisfies } \tau_t \mid R_t,S_{t+1})-\boldsymbol{\lambda}^{\top}\mathbf{C}_t.
\end{equation}
The success term measures whether $P_t$ implements $\tau_t$ while preserving the existing functionality of $R_t$ and satisfying repository-specific requirements. The cost vector $\mathbf{C}_t$ captures token consumption, execution time, verification effort, and skill maintenance overhead. The weight vector $\boldsymbol{\lambda}$ assigns nonnegative penalties to these costs, balancing implementation success and efficiency.

The PR-guided optimization objective is then defined as
\begin{equation}
\label{eq:objective}
\max J=\mathbb E\left[\sum_{t=1}^{T} U_t(P_t,R_t,\tau_t,S_{t+1})\right].
\end{equation}
Maximizing $J$ encourages the system to improve implementation success and efficiency across successive requests through continuous skill evolution.

The cost vector $\mathbf{C}_t$ in Equation~\eqref{eq:utility} captures the computational and operational overhead of the entire PR-guided development process:
\begin{equation}
\label{eq:cost}
\mathbf{C}_t=(C_t^{\rm detect},C_t^{\rm construct},C_t^{\rm verify},C_t^{\rm skill},C_t^{\rm execute}).
\end{equation}
Here, $C_t^{\rm detect}$ covers historical PR retrieval, $C_t^{\rm construct}$ covers candidate repair and synthesis, $C_t^{\rm verify}$ covers testing, $C_t^{\rm skill}$ covers skill updates, and $C_t^{\rm execute}$ covers PR generation. Each cost includes computational resources, tokens, and execution time, weighted by $\boldsymbol{\lambda}$. We report tokens and wall time separately.

The optimization objective in Equation~\eqref{eq:objective} is subject to correctness, safety, and resource constraints:
\begin{equation}
\label{eq:constraints}
\begin{aligned}
&\Pr(P_t \text{ satisfies } \tau_t \mid R_t,S_{t+1}) \geq \delta,\\
&\Pr(R_{t+1} \text{ preserves } R_t) \geq \gamma,
\end{aligned}
\qquad
\mathbf{C}_t \preceq \mathbf{B}.
\end{equation}
Here, $\delta$ and $\gamma$ specify minimum requirements for task completion and preservation of existing functionality, while $\mathbf{B}$ defines the resource budget. These constraints define the feasible region of the optimization problem, with probabilities representing desired reliability rather than guarantees from available tests. The staged process approximates this constrained objective without updating model weights or guaranteeing global optimality.

\section{Method: From PR Records to Aligned Skills}
\label{sec:method}
\ours uses each incoming request to both improve the repository and evolve reusable skills. Figure~\ref{fig:story} illustrates the sequential workflow. The system initializes a PR-record index and skill store from the repository, available historical PRs, and optional seed skills. Each round diagnoses the request, constructs and verifies PR records, evolves skills, implements and verifies the PR, and records feedback.

\subsection{PR Record Construction}

Given repository $R_t$ and request $\tau_t$, the system identifies relevant functionality, code, dependencies, and maintenance needs through feature localization~\cite{li2026featx,xia2025agentless}. It retrieves historical PRs $P_{<t}$ based on functional similarity and code overlap, and reconstructs additional PRs by segmenting historical repository commits into coherent changes. These PRs are classified as usable, repairable, outdated, irrelevant, or unsafe based on their compatibility with $R_t$, API changes, and review or CI feedback. The system then constructs candidate PRs by reusing applicable implementations, repairing outdated changes, or synthesizing new PRs with alternative implementations and supporting tests. Each candidate is evaluated through implementation checks, tests, and repository-specific requirements. Candidates satisfying the constraints in Equation~\eqref{eq:constraints} are retained as successful PR records for skill learning, while failed candidates provide failure records through their identified errors. If no candidate passes verification, further repair or synthesis is performed within the resource budget. Unlike Socratic-SWE~\cite{xiao2026socratic}, which synthesizes software tasks, our method synthesizes PR records for skill learning.

\subsection{Distill PR Information into Skills}

Following SkillLens~\cite{yu2026skilllens}, the system converts PR information into reusable repository-specific skills using a fixed-weight LLM writer and predefined skill templates. Given PR $P_t$ and repository $R_t$, the writer extracts implementation decisions and maintenance requirements from code changes and available review or CI feedback. Each skill specifies when it applies, how to implement the change, and how to validate the result. Failed implementations and review corrections provide negative constraints and repair strategies. The skill generation process is formulated as
\begin{equation}
\label{eq:skill-induction}
\widetilde S_{t+1}
\sim p_{\theta}(S \mid R_t, P_{<t}, \tau_t).
\end{equation}
Here, $p_{\theta}$ denotes the fixed-weight LLM writer, and $\widetilde S_{t+1}$ contains the generated candidate skills. Each skill retains its supporting PR records and repository-version information for subsequent alignment with the existing skill set. The skills span five families:
\begin{itemize}
    \item \textbf{PR Detection.} Assesses relevance, freshness, and quality to identify usable, repairable, outdated, or unsafe PRs.
    \item \textbf{PR Construction.} Guides task decomposition, PR reuse, repair, and synthesis to generate diverse, verifiable candidates.
    \item \textbf{Feature Implementation.} Captures implementation patterns, abstraction alignment, and cross-file dependencies.
    \item \textbf{Maintenance Completion.} Specifies testing, documentation, configuration, logging, compatibility, and release requirements.
    \item \textbf{Skill Evolution.} Uses successful and failed PR records to guide skill updates, specialization, conflict resolution, and supersession.
\end{itemize}

\begin{table}[!ht]
\caption{\textbf{Skill alignment operations.} Actions determine how new PR information updates the existing skill set.}
\label{tab:operations}
\centering\small
\begin{tabularx}{\columnwidth}{lY}
\toprule
Operation & Decision and state change \\
\midrule
\textsc{Add} & Create a new skill when no applicable skill exists. \\
\textsc{Update} & Revise an existing skill using new information. \\
\textsc{Specialize} & Narrow a skill's trigger for a specific task or context. \\
\textsc{Extend} & Add compatible procedures or maintenance requirements. \\
\textsc{Conflict} & Identify incompatible rules requiring resolution. \\
\textsc{Supersede} & Replace outdated rules while preserving their history. \\
\textsc{Keep} & Retain an existing skill when no changes are needed. \\
\bottomrule
\end{tabularx}
\end{table}

\subsection{Align New and Existing Skills}

The system aligns candidate skills $\widetilde S_{t+1}$ with the existing skill set $S_t$ to produce $S_{t+1}$, following Equation~\eqref{eq:update}. For each candidate, it identifies related skills based on task triggers, implementation procedures, repository components, and version compatibility. The system then determines whether to reuse, modify, or replace existing knowledge using seven alignment operations, as summarized in Table~\ref{tab:operations}. These operations preserve valid knowledge while incorporating new  patterns and maintenance requirements from $P_t$.

Alignment maintains the relationships between skills and their supporting PR records. Each skill records its dependencies, applicable repository versions, and related skills, allowing the system to identify reusable knowledge and resolve conflicting rules. For example, a provider PR introduces a registry-integration skill, which a subsequent exporter PR can reuse and extend with additional validation requirements. When a later PR migrates the registry, outdated registration procedures are superseded, while valid testing and documentation practices are preserved. In this way, the skill set $S_{t+1}$ evolves with repository $R_{t+1}$ while retaining useful historical knowledge. This process builds on skill revision~\cite{li2026commit}, skill bank management~\cite{li2026codeskill}, and hierarchical skill rewriting~\cite{yu2026skilllens}.
\subsection{Skill-Guided Implementation and Pool Maintenance}

Given the aligned skill set $S_{t+1}$, the system retrieves relevant skills based on task similarity, repository compatibility, and historical effectiveness. The selected skills are injected into the fixed-weight actor $A_\theta$ as repository-specific instructions, guiding code modifications and required maintenance changes. The actor generates PR $P_t$ following Equation~\eqref{eq:execution}, including applicable tests, documentation, and compatibility updates. The generated PR is verified against the correctness and safety constraints in Equation~\eqref{eq:constraints}. Failed implementations undergo bounded repair and re-verification, while accepted changes produce $R_{t+1}$. This process aims to improve the success probability in Equation~\eqref{eq:utility} while controlling execution costs, contributing to the overall objective $\max J$ in Equation~\eqref{eq:objective}.

Skill injection is modeled as
\begin{equation}
\label{eq:skill-retrieval}
S_t^{\rm use}\sim p_\theta(S\mid R_t,\tau_t,S_{t+1}),\quad |S_t^{\rm use}|\leq B_s.
\end{equation}
Here, $S_t^{\rm use}$ contains the skills injected into the actor, and $B_s$ bounds the number of retrieved skills. The system maintains the active skill pool through the alignment operations in Table~\ref{tab:operations}. New skills are added, existing skills are revised, and superseded versions are archived with their supporting PR records. Only active skills are eligible for retrieval, preventing outdated versions from accumulating while preserving historical knowledge. The final PR, verification results, and skill usage are recorded for subsequent skill evolution.

Algorithm~\ref{alg:skill-dreaming} summarizes the workflow of \ours. For each request $\tau_t$, the method constructs and verifies PR records, distills reusable skills, and aligns them with $S_t$ to obtain $S_{t+1}$. The updated skills guide feature implementation and maintenance, producing a PR $P_t$ that is verified before updating $R_t$. Its outcomes inform subsequent skill evolution.

\begin{algorithm}[!ht]
\caption{\ours Algorithm}
\label{alg:skill-dreaming}
\small
\begin{algorithmic}[1]
\Require Repository $R_0$, historical PRs $P_{<t}$,
initial skills $S_0$, requests $\{\tau_t\}_{t=0}^{T-1}$
\Ensure Evolved repository $R_T$ and skill set $S_T$

\For{$t=0,\ldots,T$}
    \State \textbf{PR Record Construction}
    \State Localize features; assess PR relevance, freshness, and quality
    \State Construct diverse candidates via reuse, repair, and synthesis
    \State Verify candidates using tests, review, and failure records

    \State \textbf{Skill Generation and Alignment}
    \State Generate $\widetilde S_{t+1}$ using Eq.~\eqref{eq:skill-induction}
    \State Update $S_{t+1}$ using Eq.~\eqref{eq:update}
    \State Apply Add, Update, Specialize, Extend, Conflict,
    \Statex \hspace{\algorithmicindent} Supersede, or Keep; maintain skill provenance

    \State \textbf{Skill-Guided Repository Implementation}
    \State Retrieve $S_t^{\rm use}$ using Eq.~\eqref{eq:skill-retrieval}
    \State Implement features with abstraction and cross-file alignment
    \State Complete tests, docs, config, logging, examples,
    \Statex \hspace{\algorithmicindent} compatibility, and release obligations
    \State Verify changes and record outcomes for subsequent rounds
    \State Generate $P_t$ and update $R_{t+1}$ using Eq.~\eqref{eq:execution}

\EndFor

\State \Return $R_T,S_T$
\end{algorithmic}
\end{algorithm}

\section{Experimental Design}
\label{sec:experiments}

\subsection{Benchmarks and Baselines}

We evaluate \ours on SWE-bench Pro~\cite{deng2025swebenchpro} and SWE-bench Verified~\cite{jimenez2024swebench,openai2024verified}. The Pro subset includes tasks from Ansible, qutebrowser, and Open Library. The evaluated Verified instances cover 12 repositories. Each task uses its historical base commit. Only previously available PR information and skills are used to prevent future information leakage.

We compare \ours with traditional retrieval methods, including BM25~\cite{robertson2009bm25}, dense retrieval~\cite{openai2024embeddings}, and hybrid retrieval. We also compare with recent skill-learning methods, including Trace2Skill~\cite{ni2026trace2skill}, Learning to Commit~\cite{li2026commit}, SkillLens~\cite{yu2026skilllens}, CODESKILL~\cite{li2026codeskill}, and Socratic-SWE~\cite{xiao2026socratic}. Additional ablations examine no-skill settings, append-only updates, positive-only skills, stale skill filtering, and alternative rewriting strategies.

Experiments use Claude Opus-5 and Sonnet-4.6 through the Anthropic SDK~\cite{anthropic2026sdk}. PR metadata is collected through GitHub APIs~\cite{github2026pulls}. We use mini-swe-agent~\cite{mini2025agent} and official SWE-bench evaluation procedures~\cite{jimenez2024swebench,deng2025swebenchpro,openai2024verified}, with Docker for execution~\cite{docker2026docs}. Benchmark retrieval uses dense cosine similarity with top eight results and a threshold of 0.20. Dense embeddings are generated using \texttt{text-embedding-3-small}~\cite{openai2024embeddings}.

\subsection{Metrics and Statistical Interpretation}

We evaluate generated PRs $P_t$ using four metrics. \textbf{Official resolved rate} ($R_{\rm official}=N_{\rm resolved}/N$) measures tasks resolved by the official benchmark harness. \textbf{Task Success (TS)} ($\mathrm{TS}=\Pr(\text{task succeeds})$) measures the probability of successful task completion under the corresponding experimental setting. \textbf{Safe rate} ($R_{\rm safe}=N_{\rm safe}/N$) measures PRs satisfying the safety requirements in Equation~\eqref{eq:constraints}, with valid, nonempty test execution. It does not guarantee feature completion. \textbf{Maintenance Surface F1} measures whether skill-guided implementation addresses all required repository changes, rather than only the core functionality. Let $G_t$ and $C(P_t)$ denote the required and implemented change categories, excluding \texttt{other}. We compute
\begin{equation}
\label{eq:surface}
p_t=\frac{|C(P_t)\cap G_t|}{|C(P_t)|},\qquad
r_t=\frac{|C(P_t)\cap G_t|}{|G_t|},\qquad
f_t=\frac{2p_tr_t}{p_t+r_t}.
\end{equation}
We report the mean $f_t$ across tasks. Categories include code, tests, documentation, configuration, examples, and changelogs. Per-category recall considers only tasks requiring that category. This metric evaluates whether skill-guided implementation addresses all necessary repository components, but does not verify the correctness of individual changes. Logging and compatibility are not evaluated separately.

\textbf{Localization.} Let $G$ denote the gold locations and $T_k$ the top-$k$ retrieved locations. Top-$k$ Hit Rate measures the probability of retrieving at least one correct location, while Recall@$k$ measures the expected fraction of gold locations retrieved. We compute
\begin{equation}
\mathrm{Hit@}k=\Pr(T_k\cap G\neq\emptyset),\quad \mathrm{Recall@}k=\mathbb{E}\left[\frac{|T_k\cap G|}{|G|}\right].
\end{equation}
Mean Reciprocal Rank (MRR) measures the average reciprocal rank of the first correct location, assigning zero when no correct location is retrieved. All metrics are evaluated on tasks with nonempty gold locations.

\textbf{Efficiency.} We measure efficiency using average token consumption and execution time per task. Token consumption includes both input and output tokens, while execution time measures wall-clock time in seconds. Offline skill construction costs are reported separately when excluded.

\section{Results}
\label{sec:results}

\subsection{End-to-End Evaluation}
\label{sec:endtoend}

\begin{table}[!ht]
\caption{\textbf{Official resolution rates on SWE-bench Pro and SWE-bench Verified.} Resolved rates are evaluated using the official benchmark harness. Base denotes the default mini-agent without skills, and $\Delta$ represents the improvement achieved by Ours over Base.}
\label{tab:benchmarks}
\centering\small
\setlength{\tabcolsep}{5pt}
\resizebox{\linewidth}{!}{%
\begin{tabular}{llrrr}
\toprule
Benchmark & Backbone & Base & Ours & $\Delta$ \\
\midrule
SWE-bench Pro~\cite{deng2025swebenchpro} & Claude Opus 5 & 0.79 & \cellcolor{sdshade}\textbf{0.88} & \cellcolor{sdshade}\textbf{+9.0\%} \\
SWE-bench Verified~\cite{openai2024verified} & Claude Opus 5 & 0.83 & \cellcolor{sdshade}\textbf{0.92} & \cellcolor{sdshade}\textbf{+9.0\%} \\
SWE-bench Verified~\cite{openai2024verified} & Claude Sonnet 4.6 & 0.77 & \cellcolor{sdshade}\textbf{0.79} & \cellcolor{sdshade}\textbf{+2.0\%} \\
\bottomrule
\end{tabular}%
}
\end{table}

\begin{table*}[!ht]
\caption{\textbf{Main results with Claude Opus 5 on SWE-bench Verified.} \textbf{Base Agent} uses the built-in mini-agent without skills; \textbf{Append-only} adds new skills without updating existing ones. \textbf{Resolved Rate} is the fraction of tasks resolved by the official benchmark harness; \textbf{Safe Rate} ($R_{\rm safe}$) is the fraction of PRs satisfying the safety requirements in Eq.~\eqref{eq:constraints}; \textbf{Existing} is the historical existing-test pass rate; \textbf{Surf. F1} is the mean Maintenance Surface F1 defined in Eq.~\eqref{eq:surface}; \textbf{Algn.} measures historical alignment. Best results are shown in bold.}
\label{tab:click-all}
\centering\small
\setlength{\tabcolsep}{8pt}
\resizebox{0.7\textwidth}{!}{%
\begin{tabular}{lccccc}
\toprule
Method & Resolved Rate $\uparrow$ & Safe Rate $\uparrow$ & Existing $\uparrow$ & Surf. F1 $\uparrow$ & Algn. $\uparrow$ \\
\midrule
Base Agent & 0.83 & 0.83 & 0.83 & 0.62 & 0.38 \\
\midrule
Append-only & 0.88 & 0.94 & 0.94 & 0.72 & 0.44 \\
Trace2Skill~\cite{ni2026trace2skill} & 0.76 & 0.78 & 0.78 & 0.60 & 0.37 \\
Learning to Commit~\cite{li2026commit} & 0.72 & 0.67 & 0.67 & 0.64 & \textbf{0.49} \\
SkillLens~\cite{yu2026skilllens} & 0.90 & 0.61 & 0.61 & 0.59 & 0.37 \\
CODESKILL~\cite{li2026codeskill} & 0.79 & 0.56 & 0.56 & 0.46 & 0.32 \\
Socratic-SWE~\cite{xiao2026socratic} & 0.91 & 0.72 & 0.72 & 0.64 & 0.45 \\
\midrule
\rowcolor{sdshade}
\ours (Ours) & \textbf{0.92} & \textbf{1.00} & \textbf{1.00} & \textbf{0.74} & 0.44 \\
\bottomrule
\end{tabular}%
}
\end{table*}

Table~\ref{tab:benchmarks} shows that \ours consistently improves task resolution across benchmarks and backbone models. With Claude Opus 5, it achieves 88\% on SWE-bench Pro and 92\% on SWE-bench Verified, improving the base agent by 9\% on both benchmarks. The gains also extend to Claude Sonnet 4.6, demonstrating effectiveness across different model capabilities. We exclude Sonnet 4.6 from SWE-bench Pro due to its weak baseline performance.

To further evaluate \ours, we compare it with existing skill learning methods in Table~\ref{tab:click-all}. \ours achieves the highest resolved rate of 92\%, outperforming all baselines, including append-only skill accumulation. It also achieves a perfect Safe Rate of 1.00 and the highest Maintenance Surface F1 of 0.74, demonstrating stronger preservation of existing functionality and more comprehensive maintenance of related repository components. These results highlight the advantages of continuously evolving repository-specific skills over static or append-only approaches.

\subsection{Cross-Repository Generalization}
\label{sec:generalization}

\begin{table}[!ht]
\caption{\textbf{Cross-repository generalization with Claude Opus 5.}
Task Success (TS) measures task completion, and Surf. F1 denotes the Maintenance Surface F1 defined in Eq.~\eqref{eq:surface}.}
\label{tab:cross-repo}
\centering\small
\setlength{\tabcolsep}{3pt}
\resizebox{0.9\columnwidth}{!}{%
\begin{tabular}{llcccc}
\toprule
Repository & Category & Base TS & Ours TS & Base F1 & Ours F1 \\
\midrule
Click & CLI / Dev Tools & 0.72 & \cellcolor{sdshade}\textbf{0.83} & 0.70 & \cellcolor{sdshade}\textbf{0.72} \\
Black & CLI / Formatter & 0.53 & \cellcolor{sdshade}\textbf{0.60} & 0.48 & \cellcolor{sdshade}\textbf{0.67} \\
Rich & CLI / Rendering & 0.90 & \cellcolor{sdshade}\textbf{1.00} & 0.78 & \cellcolor{sdshade}\textbf{0.90} \\
\midrule
HTTPX & Web / HTTP & 0.80 & \cellcolor{sdshade}\textbf{0.93} & 0.72 & \cellcolor{sdshade}\textbf{0.80} \\
Arrow & Data / Scientific & 0.58 & \cellcolor{sdshade}\textbf{0.75} & 0.71 & \cellcolor{sdshade}\textbf{0.74} \\
Agents & Agent Framework & 0.50 & \cellcolor{sdshade}\textbf{0.70} & 0.85 & \cellcolor{sdshade}\textbf{0.88} \\
\bottomrule
\end{tabular}%
}
\end{table}

We evaluate cross-repository generalization on six repositories spanning four software domains: CLI and development tools (\href{https://github.com/pallets/click}{Click}, \href{https://github.com/psf/black}{Black}, and \href{https://github.com/Textualize/rich}{Rich}), web and HTTP libraries (\href{https://github.com/encode/httpx}{HTTPX}), data and scientific computing (\href{https://github.com/arrow-py/arrow}{Arrow}), and agent frameworks (\href{https://github.com/openai/openai-agents-python}{Agents}). For each repository, we first split historical PRs into skill-construction and evaluation sets. \ours then constructs repository-specific skills from an initially empty skill pool using historical PR data and applies the evolving skills to evaluation PRs. As shown in Table~\ref{tab:cross-repo}, \ours improves task success across all six repositories, with gains ranging from 0.07 to 0.20, while also improving Maintenance Surface F1. These results demonstrate the ability of \ours to adapt to diverse repository structures and development conventions.

\begin{table}[!ht]
\caption{\textbf{Code localization at different granularities.} Top-$k$ indicates whether a correct location is retrieved, R@$k$ measures the fraction of correct locations retrieved, and MRR measures the reciprocal rank of the first correct result.}
\label{tab:localization}
\centering\small
\setlength{\tabcolsep}{3pt}
\resizebox{0.7\columnwidth}{!}{%
\begin{tabular}{lccccc}
\toprule
Granularity & Top-1 & Top-3 & R@5 & R@10 & MRR \\
\midrule
Module & \cellcolor{sdshade}\textbf{1.00} & \cellcolor{sdshade}\textbf{1.00} & \cellcolor{sdshade}\textbf{0.99} & \cellcolor{sdshade}\textbf{0.99} & \cellcolor{sdshade}\textbf{1.00} \\
File & 0.64 & 0.90 & 0.92 & 0.96 & 0.77 \\
Function & 0.25 & 0.39 & 0.33 & 0.33 & 0.33 \\
\bottomrule
\end{tabular}%
}
\end{table}

\begin{table}[!ht]
\caption{\textbf{PR quality classification results.} Precision, recall, and F1 for SWE-bench Pro ansible dataset of 1,000 PRs.}
\label{tab:assessment-scores}
\centering\small
\setlength{\tabcolsep}{3pt}
\resizebox{0.9\columnwidth}{!}{%
\begin{tabularx}{\columnwidth}{Yrrrr}
\toprule
PR Category & Precision & Recall & F1 & Support \\
\midrule
Usable & 0.93 & 0.94 & 0.94 & 260 \\
Repairable & 0.87 & 0.90 & 0.88 & 50 \\
Outdated & 0.90 & 0.91 & 0.90 & 190 \\
Irrelevant & \cellcolor{sdshade}\textbf{0.96} & \cellcolor{sdshade}\textbf{0.93} & \cellcolor{sdshade}\textbf{0.94} & 450 \\
Unsafe & 0.85 & 0.88 & 0.86 & 50 \\
\bottomrule
\end{tabularx}
}
\end{table}

\begin{table}[!ht]
\caption{\textbf{Comparison of PR implementation methods.} Task Success (TS) measures successful task completion, and Maintenance Surface F1 evaluates the coverage and precision of required repository changes.}
\label{tab:candidates}
\centering\small
\setlength{\tabcolsep}{4pt}
\resizebox{0.8\columnwidth}{!}{%
\begin{tabular}{lcc}
\toprule
Method & TS & Surface F1 \\
\midrule
Single Sample & 0.94 & 0.80 \\
Best-of-$K$ & 1.00 & 0.87 \\
Trace2Skill~\cite{ni2026trace2skill} & 0.78 & 0.60 \\
SkillLens-inspired composition~\cite{yu2026skilllens} & 0.61 & 0.71 \\
\midrule
\cellcolor{sdshade}\textbf{\ours} &
\cellcolor{sdshade}\textbf{1.00} &
\cellcolor{sdshade}\textbf{0.87} \\
\bottomrule
\end{tabular}%
}
\end{table}

\begin{table}[!ht]
\caption{\textbf{Impact of information sources for skill learning.} Comparison of task success (TS), skill reuse precision, and recall across different information sources and methods.}
\label{tab:sources}
\centering\small
\setlength{\tabcolsep}{4pt}
\resizebox{0.9\columnwidth}{!}{%
\begin{tabular}{lccc}
\toprule
Information Source / Method & TS & Reuse Precision & Recall \\
\midrule
Positive PRs only & 0.84 & 0.67 & 0.78 \\
Review / reply only & 0.83 & 0.48 & 0.56 \\
Trace2Skill~\cite{ni2026trace2skill} & 0.72 & 0.61 & 0.76 \\
Socratic~\cite{xiao2026socratic} & 0.83 & 0.67 & 0.77 \\
\midrule
\cellcolor{sdshade}\textbf{\ours} &
\cellcolor{sdshade}\textbf{0.89} &
\cellcolor{sdshade}\textbf{0.72} &
\cellcolor{sdshade}\textbf{0.82} \\
\bottomrule
\end{tabular}%
}
\end{table}

\begin{table*}[!ht]
\caption{\textbf{Impact of skill relation management.} Comparison of skill filtering strategies across two settings. TS denotes Task Success, and Surface F1 denotes Maintenance Surface F1.}
\label{tab:relations}
\centering\small
\setlength{\tabcolsep}{5pt}
\resizebox{0.7\textwidth}{!}{%
\begin{tabular}{lcccccc}
\toprule
& \multicolumn{3}{c}{Skill Filtering} & \multicolumn{3}{c}{Hybrid Retrieval} \\
\cmidrule(lr){2-4}\cmidrule(lr){5-7}
Method & TS & Surface F1 & Precision & TS & Surface F1 & Precision \\
\midrule
No Supersession & 0.67 & 0.51 & 0.92 & 0.83 & 0.74 & 0.97 \\
No Stale Filtering & 0.72 & 0.64 & \textbf{0.95} & 0.78 & 0.71 & 0.92 \\
Neither Filter & 0.78 & 0.69 & 0.93 & 0.89 & 0.74 & \textbf{0.97} \\
Append-only & 0.89 & 0.75 & 0.92 & 0.89 & 0.74 & 0.92 \\
\midrule
\cellcolor{sdshade}\textbf{\ours (Full)} &
\cellcolor{sdshade}\textbf{0.93} &
\cellcolor{sdshade}\textbf{0.75} &
\cellcolor{sdshade} 0.93 &
\cellcolor{sdshade}\textbf{0.94} &
\cellcolor{sdshade}\textbf{0.77} &
\cellcolor{sdshade}0.90 \\
\bottomrule
\end{tabular}%
}
\end{table*}

\begin{table}[!ht]
\caption{Comparison of skill updating methods. TS denotes Task Success.}
\label{tab:rewrite}
\centering\small
\setlength{\tabcolsep}{5pt}
\resizebox{0.6\columnwidth}{!}{%
\begin{tabular}{lc}
\toprule
Skill Update Method & TS \\
\midrule
Append-only & 0.89 \\
Parent-only & 0.87 \\
Rewrite-all & 0.82 \\
SkillLens-style selective~\cite{yu2026skilllens} & 0.83 \\
No conflict/supersede & 0.78 \\
Learning to Commit-style~\cite{li2026commit} & 0.78 \\
SkillLens-style~\cite{yu2026skilllens} & 0.89 \\
\midrule
\cellcolor{sdshade}\textbf{\ours (Full)} &
\cellcolor{sdshade}\textbf{0.93} \\
\bottomrule
\end{tabular}%
}
\end{table}

\begin{table}[!ht]
\caption{\textbf{Efficiency comparison.} Comparison of task success, maintenance quality (Surface F1), token consumption (Tokens), and execution time.}
\label{tab:budget}
\centering\small
\setlength{\tabcolsep}{5pt}
\resizebox{0.8\columnwidth}{!}{%
\begin{tabular}{lcccc}
\toprule
Method & TS & Surface F1 & Tokens & Time (s) \\
\midrule
SkillLens~\cite{yu2026skilllens} & 0.61 & 0.71 & \textbf{14,300} & \textbf{32.90} \\
\cellcolor{sdshade}\textbf{\ours} &
\cellcolor{sdshade}\textbf{0.93} &
\cellcolor{sdshade}\textbf{0.80} &
\cellcolor{sdshade}26,894 &
\cellcolor{sdshade}36.00 \\
\bottomrule
\end{tabular}%
}
\end{table}

\subsection{PR Record Analysis and Implementation Evaluation}

Table~\ref{tab:localization} evaluates code localization at module, file, and function levels. Localization performs best at the module level, followed by file-level retrieval, while function-level localization remains more challenging.

Building on localization, Table~\ref{tab:assessment-scores} illustrates PR quality classification across five categories using a set of 1,000 SWE-bench Pro Ansible PRs. All categories achieve F1 scores above 0.86, with irrelevant PRs reaching 0.945.

Finally, Table~\ref{tab:candidates} compares PR implementation methods. Single Sample generates one patch, while Best-of-$K$ selects from multiple candidates. Trace2Skill and SkillLens composition serve as skill-based baselines. \ours achieves a Task Success of 1.000 and a Maintenance Surface F1 of 0.871, outperforming Single Sample and both skill-based baselines while matching Best-of-$K$ on shared candidate outputs.

Table~\ref{tab:sources} examines how different historical PR information sources affect the quality of learned repository skills and their effectiveness in implementing future PRs. We compare skills constructed from successful PRs, review discussions, and multiple information sources against existing skill-learning methods. \ours achieves the highest Task Success of 0.89, outperforming positive-only skills, review-only skills, Trace2Skill, and the Socratic baseline.

\subsection{Ablation Studies}

Table~\ref{tab:relations} presents an ablation study of skill relation management under two settings: Skill Filtering, which directly evaluates relation-based filtering, and Hybrid Retrieval, which applies filtering after combined retrieval. We ablate supersession filtering, stale-skill filtering, and both, with append-only as an additional baseline. The full \ours achieves the highest TS of 0.93 and 0.94, with Surface F1 scores of 0.75 and 0.77, respectively. These results demonstrate the importance of jointly maintaining supersession and stale-skill filtering.

Following the ablation of skill relation management, Table~\ref{tab:rewrite} further investigates the impact of different skill updating strategies. Append-only adds new skills without modifying existing ones. Parent-only updates the directly related parent skill, while Rewrite-all rewrites all relevant skills. SkillLens-style selection~\cite{yu2026skilllens} selectively revises existing skills based on their relevance to new information. No conflict/supersede removes explicit conflict resolution and supersession handling. Learning to Commit-style~\cite{li2026commit} extracts reusable development knowledge from commit histories, while SkillLens-style~\cite{yu2026skilllens} organizes and reuses learned skills. In contrast, \ours updates repository skills while maintaining their relationships and validity. The full \ours achieves a TS of 0.93, outperforming all skill updating variants, while existing PR retrieval achieves 1.00.

Finally, we examine the impact of skill filtering through an additional ablation. Disabling filtering reduces Maintenance Surface F1 from 0.867 to 0.756, a decrease of 0.111. This result further highlights the importance of filtering outdated and superseded skills during repository evolution.

\subsection{Continuous Skill Evolution}

\begin{figure}[!ht]
\centering
\includegraphics[width=0.9\columnwidth]{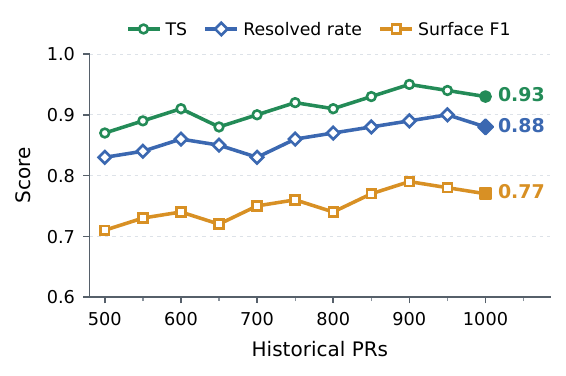}
\caption{\textbf{Continuous skill evolution through PR streams.}
TS, resolved rate, and Surface F1 of \ours with Claude Opus 5
on SWE-bench Pro as new PRs arrive and repository-specific
skills are incrementally updated.}
\label{fig:pr-scaling}
\end{figure}

We investigate whether \ours can continuously improve repository-specific skills as new PRs arrive. Using Claude Opus 5 under the SWE-bench Pro headline setting, we initialize the skill set with 500 historical PRs and incrementally update it with 50 additional PRs at each stage, reaching 1,000 PRs. Figure~\ref{fig:pr-scaling} illustrates the evolution of task performance and maintenance quality. All three metrics exhibit an overall upward trend despite occasional fluctuations, suggesting that accumulated PR knowledge can progressively improve skill-guided implementation. This experiment evaluates whether \ours can adapt to changing repository conventions through continuous skill evolution.

\subsection{Efficiency Analysis}

Table~\ref{tab:budget} compares the efficiency of \ours and SkillLens in terms of task performance, maintenance quality, and computational cost. \ours achieves a TS of 0.93 and a Surface F1 of 0.80, outperforming SkillLens, which obtains 0.61 and 0.71, respectively. This improvement comes with increased resource consumption, as \ours uses 26,894 tokens and requires 36.00 seconds per task, compared with 14,300 tokens and 32.90 seconds for SkillLens. These results indicate that \ours provides substantial improvements in task success and repository maintenance, with higher token consumption but only a modest increase in execution time.

\section{Conclusion}
We propose \ours, a framework that continuously evolves repository-specific skills from pull request streams. By constructing verified PR records, distilling reusable skills, and maintaining their relationships, \ours enables coding agents to extend and maintain evolving repositories. This process transforms historical development evidence into actionable knowledge that guides future implementation decisions. As repositories evolve, \ours adapts existing skills to changing code structures, implementation conventions, and maintenance requirements. Experiments show that \ours improves the official resolved rate from 79\% to 88\% on SWE-bench Pro and from 83\% to 92\% on SWE-bench Verified. Further evaluations demonstrate its effectiveness across repositories, the importance of skill evolution and alignment, and improved task success and maintenance quality with modest additional computational costs. These findings highlight PR-driven skill evolution as a promising approach to continuous repository development and maintenance.

\bibliographystyle{ACM-Reference-Format}
\bibliography{references}
\clearpage
\appendix

\end{document}